\documentclass[conference]{ieeeconf}
\usepackage{amsmath,amssymb,graphicx}
\usepackage{ragged2e}
\usepackage[utf8]{inputenc}
\usepackage{wrapfig}
\usepackage{mathtools}
\usepackage{fixltx2e,hyperref}
\usepackage[most]{tcolorbox}
\usepackage{booktabs}

\usepackage{xcolor,colortbl}
\usepackage{changepage}
\usepackage{hhline}
\usepackage{tabularx}
\usepackage[ruled]{algorithm2e}

\begin{document}
\title{Semi-Supervised Multi-Task Learning for Lung Cancer Diagnosis}
%
\author{Naji Khosravan and Ulas Bagci\\
Email: najikh@cs.ucf.edu, bagci@crcv.ucf.edu \\
Center for Resaerch in Computer Vision (CRCV), University of Central Florida (UCF), Orlando, FL.}
%
%
%
%
%

\maketitle
\begin{abstract}
Early detection of lung nodules is of great importance in lung cancer screening. Existing research recognizes the critical role played by CAD systems in early detection and diagnosis of lung nodules. However, many CAD systems, which are used as cancer detection tools, produce a lot of false positives (FP) and require a further FP reduction step. Furthermore, guidelines for early diagnosis and treatment of lung cancer are consist of different shape and volume measurements of abnormalities. Segmentation is at the heart of our understanding of nodules morphology making it a major area of interest within the field of computer aided diagnosis systems. This study set out to test the hypothesis that joint learning of false positive (FP) nodule reduction and nodule segmentation can improve the computer aided diagnosis (CAD) systems' performance on both tasks. To support this hypothesis we propose a 3D deep multi-task CNN to tackle these two problems jointly. We tested our system on LUNA16 dataset and achieved an average dice similarity coefficient (DSC) of $\textbf{91\%}$ as segmentation accuracy and a score of nearly $\textbf{92\%}$ for FP reduction. As a proof of our hypothesis, we showed improvements of segmentation and FP reduction tasks over two baselines. Our results support that joint training of these two tasks through a multi-task learning approach improves system performance on both. We also showed that a semi-supervised approach can be used to overcome the limitation of lack of labeled data for the 3D segmentation task. 
\end{abstract}

\section{Introduction}
Lung cancer has the highest rate of mortality among the cancer related deaths \cite{CAAC:CAAC21387}. Lung nodules are primary indicators of lung cancer and early diagnosis of them can increase survival rate considerably. Detecting these nodules along with different shape and size measurements enables radiologists to have an early diagnosis of their malignancy \cite{macmahon2005guidelines}. Toward this goal, many computer aided systems has been developed and shown to play a key role in early diagnosis of lung cancer \cite{sluimer2006computer,desantis2014cancer}. 

Existing automated lung nodule detection systems produce a lot of false positives (FP). Hence, there is an additional step needed to further reduce these FPs. This is a fundamental component of nearly all available CAD systems in the literature \cite{sluimer2006computer}. In the FP reduction step, candidates are being classified as \textit{nodule} or \textit{non-nodule} using discriminative features. Segmentation, on the other hand, is of interest as it is the first step toward quantification and different shape/size and volume measurements. In this study, we argue about the use of segmentation within the FP removal step. Since a good 3D segmentation of lung nodules leads to accurate volume/shape measurement analysis in cancer screening and treatment planning, it can be used as a discriminator information for FP identification. Although some studies used different nodule attributes in a multi-task manner with pretrained networks to do nodule characterization \cite{hussein2017risk}, till now, none of previous studies used segmentation within a FP reduction jointly.


\begin{figure*}[h]
\centering
\includegraphics[width=\linewidth,height=9cm]{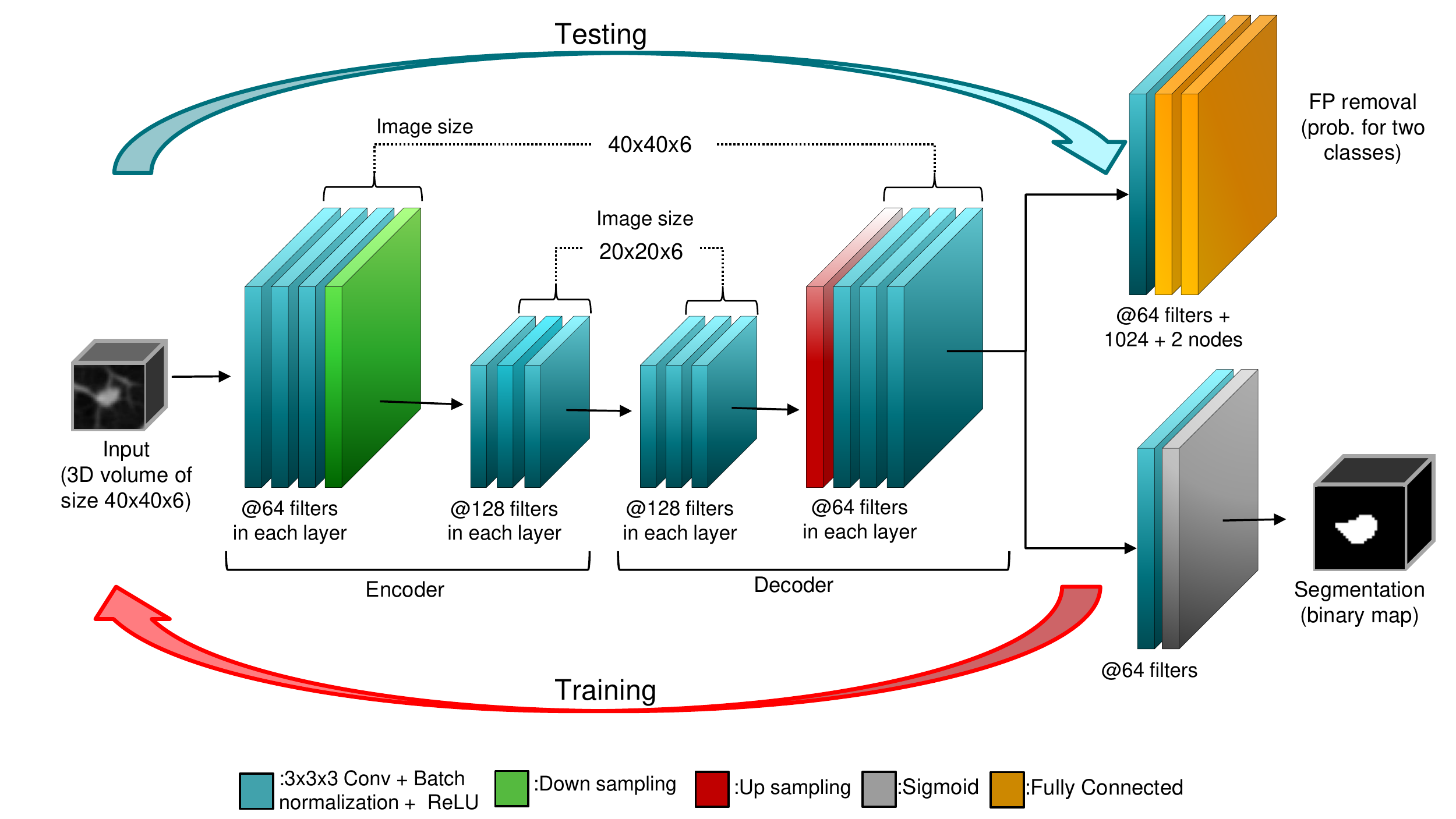}
\caption{The 3D deep multi-task CNN architecture. The size of all convolution kernels is set to $3\times 3\times 3$ with a stride of $1$ in each dimension. The downsampling and upsampling operators are performed only in the $xy$ plane. All convolution layers are $3D$. The network has $14$ shared layers, $3$ FP removal specific layers, and $2$ segmentation specific layers. Red and blue arrows show the semi-supervised learning paradigm to train the proposed network.\label{fig:architecture}}
\end{figure*}

This paper proposes a new methodology for addressing both \textit{FP reduction} and \textit{segmentation} problems, jointly. We propose a general model (Figure 1) that can perform both tasks with high accuracy through a multi-task learning (MTL) strategy. Our proposed model has a novel 3D deep encoder-decoder CNN architecture. We also exploit a semi supervised approach for training our model to avoid the need for large number of manual annotations for 3D segmentation masks. \textbf{Our contributions} can be specified as: \textbf{1)} This is the first study to propose joint segmentation and FP reduction of lung nodules through a fully 3D CNN, which is a critical step toward using CAD systems efficiently in clinical applications. \textbf{2)} Our work opens a door to possible improvements of CAD systems via a MTL approach. \textbf{3)} This work will generate fresh insight on how to tackle the problem of lack of available annotated medical image data through a semi-supervised learning method, which is more efficient if used along with MTL.

\section{Method}
The proposed 3D deep MTL algorithm is based on Convolutional Neural Network (CNN) and learns segmentation and FP reduction through some \textit{shared and task specific layers}. The proposed architecture along with the training strategy is illustrated in Fig.\ref{fig:architecture}. In the rest of this section, we explain the proposed framework step by step.

\subsection{Multi-Task Learning}
MTL allows solving multiple learning tasks at the same time by optimizing multiple loss functions instead of one \cite{caruana1998multitask}. MTL can be beneficial in multiple senses: (1) \textit{Generalization ability:} in MTL, a single model can be used to perform multiple tasks at the same time. Such as, in our case, it is desirable to have one general model, with the same accuracy if not better, instead of having multiple separate models. (2) \textit{Highlighting underlying features:} depending on the selection of the tasks, features learned from one task can act discriminative for other tasks as well. These features might not always be easy to learn by a single task network due to their complexity or more discriminating effect of other features. However, learning multiple tasks jointly can strengthen the effect of these underlying features and boost the performance on one or all tasks. (3) \textit{Dealing with lack of data:} in radiology field, it is not easy to gather large number of annotated data for training deep networks. An MTL model can benefit each task during training due to actively sharing features in relevant tasks.

The problem of jointly learning multiple tasks can be formulated as follows. Assume that we have $N$ supervised tasks. The training set for each task can be considered as $D_{n}={(x_{in},y_{in})}$. In which $i=1:k_{n}$, where $k_{n}$ is the number of training samples for the $n_{th}$ task. With $x_{in}\in X^{(n)}$ and $y_{in}\in Y^{(n)}$ the problem of learning multiple tasks, jointly, can be narrowed down to the optimization problem of:
\begin{equation}
\min_{w} \sum_{n=1}^{N} L(Y^{(n)},f(X^{(n)}) + \lambda \lVert f \rVert,
\label{equ:MTL}
\end{equation}\\
where $L: {\rm I\!R}\times {\rm I\!R} \to {\rm I\!R}^{+}$ is the loss function measuring the per-task prediction error, $f$ is the multi-task model and $w$ is the model's parameter set. In our study, we use cross entropy as loss function for both tasks. Cross entropy, also known as negative log likelihood, measures the similarity between two probability distributions and conventionally defined as:
\begin{equation}
L(Y^{(n)},f(X^{(n)})=\sum_{i=1}^{k_{n}}-y_{i}\log(f(x_{i})),
\end{equation}
where $y_{i}$s are the true labels and $f(x_{i})$s are the predictions for each task. To optimize equation \ref{equ:MTL}, ADAM optimizer was used with an initially selected learning rate of $10^{-3}$.

Since morphology (i.e., size, volume, and shape) information plays a key role in screening, diagnosis, and prognosis, we earlier postulate that this information can be effectively used for FP rejection, which is a significant challenge for most CADs. There is a strong need for reducing those findings (FPs) because it tremendously increases the workload of radiologists. We proved in the following that an MTL based CAD system can solve these two problems jointly: segmenting nodules while deciding whether they are FP or not. We believe that once the shape and appearance information can be highlighted in the shared layers of a network, other task specific layers can also learn if the nodule is a true nodule or not. In other words, features for classification and segmentation are combined through shared layers of the proposed network. To our best, this is the first study conducting this for both FP removal and segmentation.

\subsection{Architecture}
The inputs to our network are 3D volumes and the outputs are probabilities of each volume belonging to class of nodules or non-nodules. Our second output is a binary segmentation mask for those nodules. Our network has $19$ layers: the first $14$ layers are trained on both tasks, $5$ task specific layers (2 for segmentation, 3 for classification)  are trained only on one of the tasks. Each convolution layer in the architecture consists of a set of 3D convolution kernels (with size of $(3,3,3)$ and a stride of $1$) following by a batch normalization (BN) and a rectified linear unit activation (ReLU). Number of kernels in each layer is depending on its location in the architecture. A max-pooling layer with the kernel size of $(2,2)$ is used to perform down-sampling in the encoder. A bilinear interpolation is used for the up-sampling images in the decoder.

Our network \textit{forks} after  $14th$ layer into two branches (see Fig.\ref{fig:architecture}). Segmentation specific branch contains a convolution layer following by a sigmoid layer, which produces binary masks. FP reduction branch contains a convolution layer followed by two fully connected layer. The fully connected layers have $1024$ and $2$ nodes, respectively, and output the probability of each patch belonging to each one of classes (nodule vs. non-nodule).

\subsection{Semi-supervised training}
Due to the large number of parameters, deep CNNs need a large amount of annotated data to be trained efficiently. However, finding a large amount of such data is very challenging and expensive, specifically in the field of medical imaging. Semi-supervised learning methods are one way to address such issues. In semi-supervised methods, the model is initially trained on the part of data set which has labels. This model is then used to estimate labels for unlabeled data, which will be used to refine the model. The algorithm for semi-supervised learning strategy is illustrated in Algorithm \ref{alg:Semi-Supervised}. It can be argued that semi-supervised approach, if utilized naively, can lead to error propagation in the model and even cause worse performance. This problem, however, can be solved by iteratively performing prediction and training on small portions of unlabeled data and improving performance step by step. Constant improvements of results in our case supports that our algorithm perfectly handles error propagation and outperforms the baseline.
\begin{algorithm}[htb]
    \SetKwInOut{Input}{Input}
    \SetKwInOut{Output}{Output}
    \Input{labeled data: $(X_{l},Y_{l})$, unlabeled data: $X_{u}$}
    Train model $f$ on $(X_{l},Y_{l})$\\
    \For{$x$ in $X_{u}$}{
        Predict on $x \in X_{u}$\\
        Add $(x,f(x))$ to labeled data\\
        Retrain model $f$
      }
Return refined model $f$\; 
    \caption{Semi-Supervised training algorithm}
    \label{alg:Semi-Supervised}
\end{algorithm}
\section{Results}
\textbf{Data:} To evaluate our network we used Lung Nodule Analysis (LUNA16) Challenge dataset~\cite{LUNA16}. This dataset is gathered from the largest publically available LIDC-IDRI dataset. Scans with a slice thickness greater than $2.5$ mm were excluded from the dataset leaving a total of $888$ chest CT scans. The dataset contains the location of nodules accepted by at least $3$ out of $4$ radiologists leading to a total of $1186$ nodule annotations. We performed our experiments on a total number of more than $500,000$ candidate locations provided by the dataset for the FP reduction task, which are a combination of outputs of candidate generation methods in the literature. This dataset is divided into $10$ subsets by the provider. We performed $10$-fold cross validation to evaluate our method. To handle the unbalance ratio between nodules and non-nodules we performed data augmentation on the nodules (shift in $6$ directions). It should be mentioned that the number of segmentation masks available for this study was only $\textbf{270}$ out of $1186$ total nodule annotations and the masks for the rest ($916$ nodules) was created using the proposed semi-supervised strategy.

\begin{figure*}[h]
\includegraphics[scale=0.32]{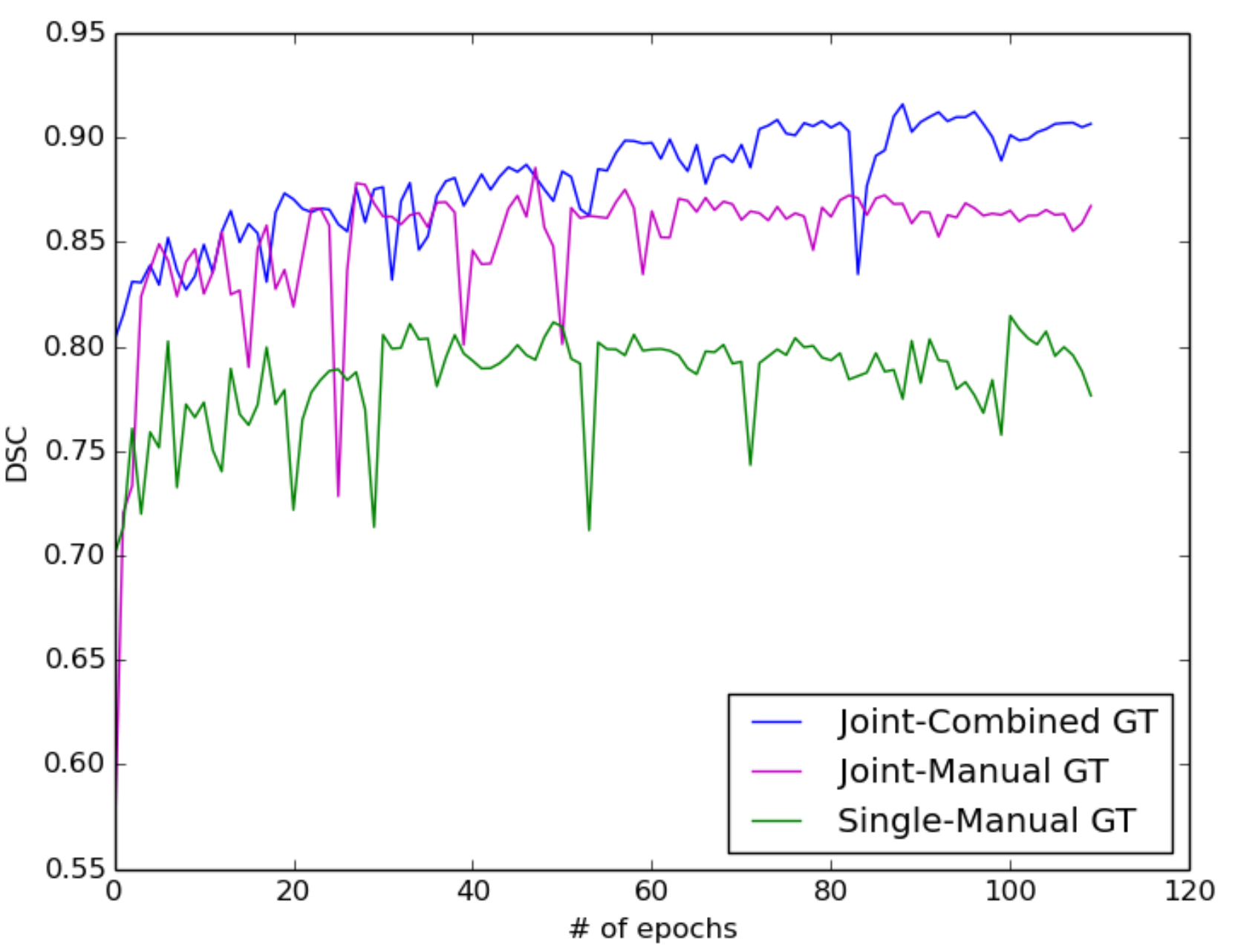}
\includegraphics[scale=0.32]{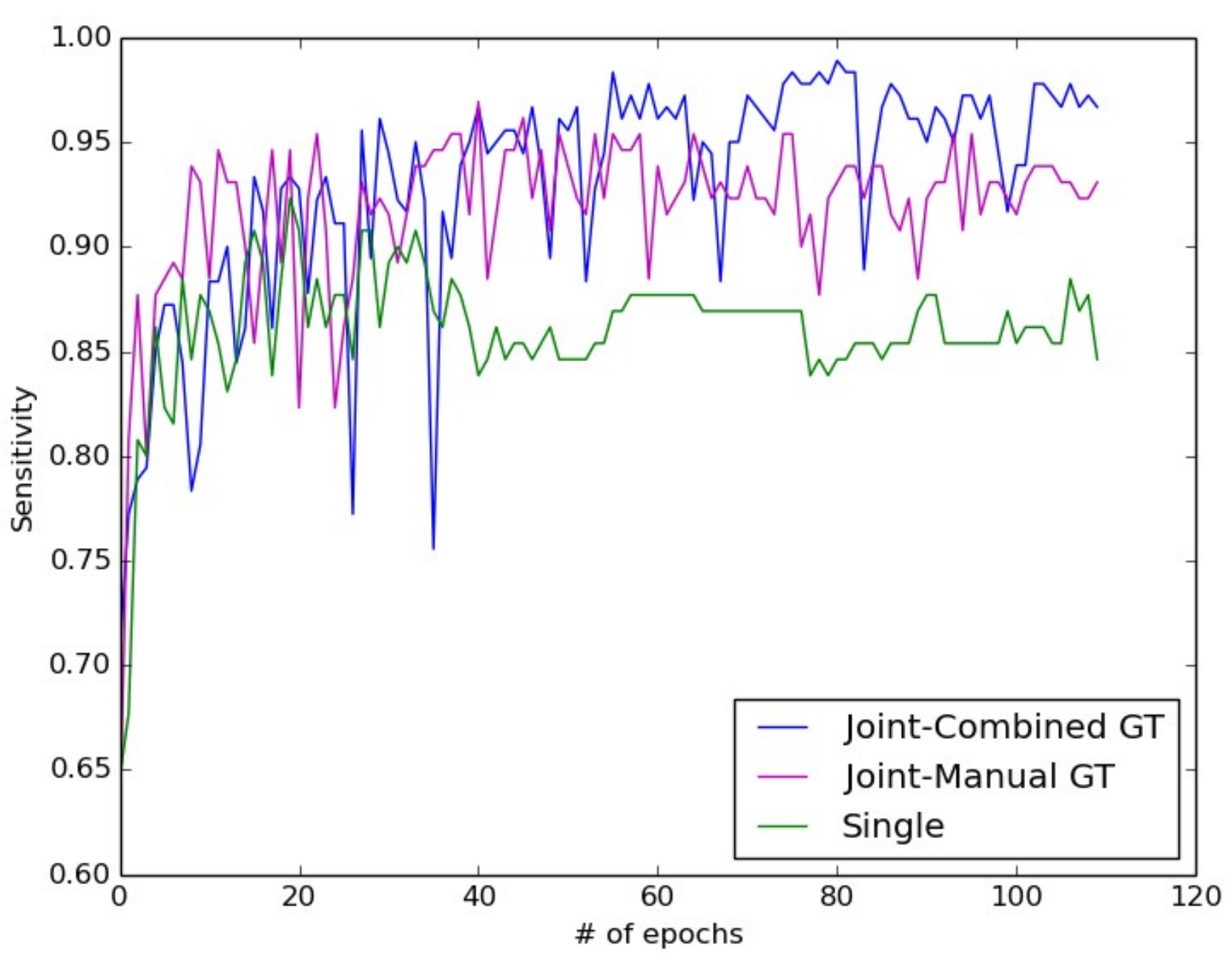}
\includegraphics[scale=0.3]{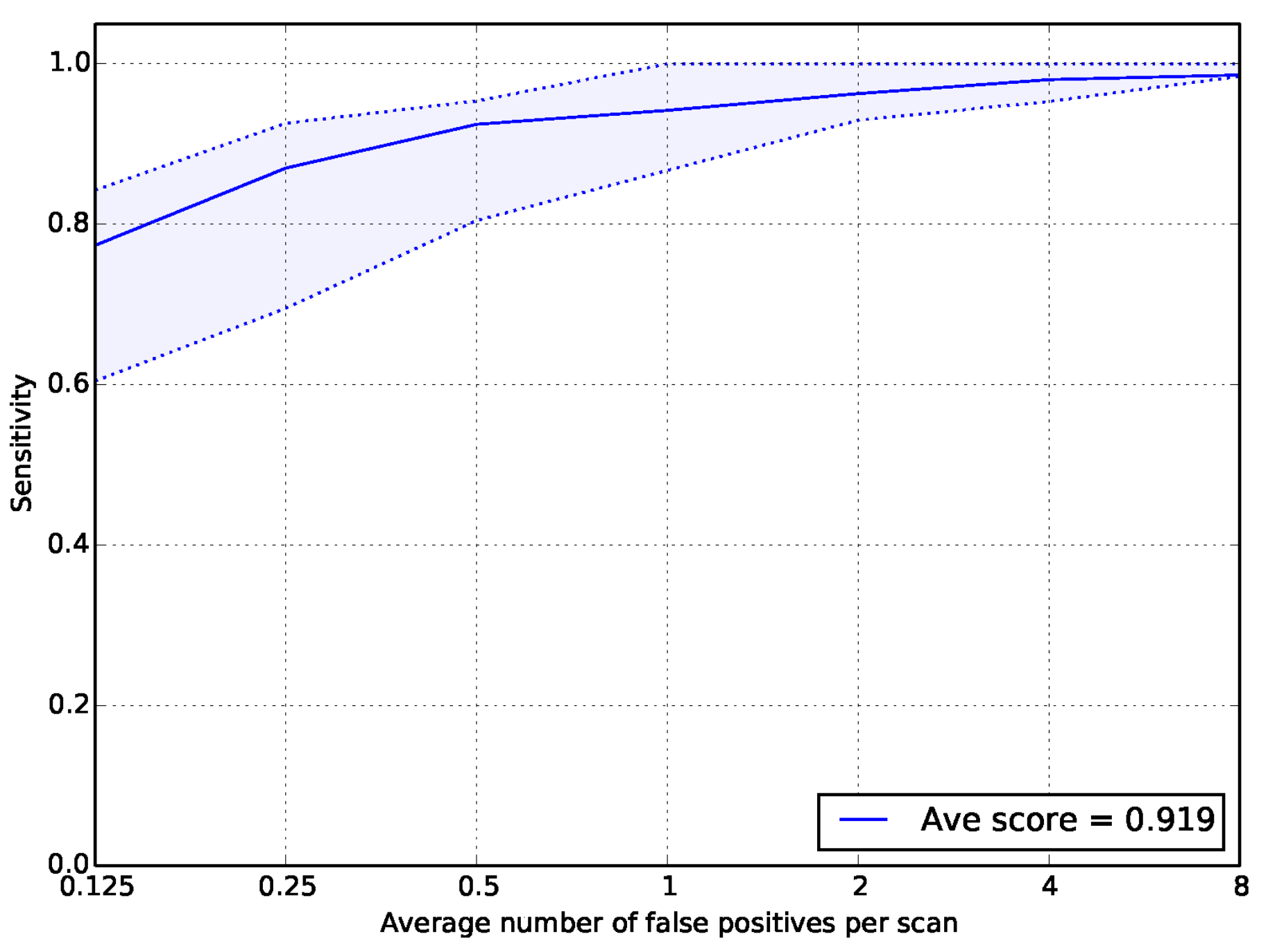}
\caption{Comparison of two baselines with proposed method. First baseline is single task network, second is semi-supervised MTL. \textit{Left}: Dice similarity coefficient over first $100$ learning epochs is shown. \textit{middle}: Showing sensitivity for FP reduction task over the first $100$ epochs. Improvement of segmentation through different training strategies are depicted. \textit{Right}: is showing the FROC curve.\label{fig:dicecomparison}}
\end{figure*}

\textbf{Segmentation:} We used Dice Similarity Coefficient (DSC) as the metric to measure segmentation accuracy. To show the improvements, we compared the final model to $2$ baselines of our model. Learning curves are plotted in Fig~\ref{fig:dicecomparison}. In first baseline, we trained the model as a single task model using only the portion of annotated data which is available (depicted as single-manual ground truth (GT) in the plot-green). In second baseline, we trained the model jointly on both segmentation and FP reduction tasks as a MTL network with the same manual GT (depicted as joint-manual GT in the plot-pink). This multi-task model was used to generate annotations for the rest of the dataset. Next, we trained the model using the semi-supervised approach (depicted as joint-combined GT in the plot-blue). Note that we trained all models from scratch.

As shown, MTL based network outperforms single task based methods and semi-supervised approach improves results of MTL further. Our network reaches a DSC of $\textbf{91\%}$ compared to the baseline which does not go beyond $\textbf{82\%}$. 

\textbf{FP reduction:}
To observe the effect of proposed semi-supervised MTL method on FP reduction performance, we compared the learning curves of three training strategies as follows. Single task network trained to only perform FP reduction (depicted as Single-green), Multi-Task using only manual GT available for the segmentation (depicted as Joint-Manual GT in the plot-pink) and Multi-Task using semi-supervised approach (depicted as Joint-Combined GT in the plot-blue). Figure~\ref{fig:dicecomparison} shows sensitivity through training epochs. As expected improvements are observed in the classification results (from $\textbf{88\%}$ to $\textbf{98\%}$). Our results also show that, beside improving segmentation results, using a semi-supervised approach benefits FP reduction task as well (Joint-Combined GT in the plot-blue). This supports our rationale behind proposing a multi-task network strongly by showing that a better segmentation, which is highlighting shape and appearance information better, helps the other relevant task (FP reduction). Summary of the best performance on each task using different learning strategies is illustrated in Table~ \ref{table:results}.

\begin{table}[htb]
\caption{Dice similarity coefficient and sensitivity for three different learning methods is shown.}
\vspace{.25cm}
\resizebox{\columnwidth}{!}{
\begin{tabular}{|l|c|c|}
\hline
\rowcolor{gray}
Training strategy & DSC\% & Sensitivity\% \\
\hline
 Single task  & 82\% & 88\%  \\
\hline
Multi task (manual GT) & 86\% & 95\% \\
\hline
Semi-Supervised multi task & \textbf{91\%} & \textbf{98\%} \\
\hline
\end{tabular}}
\label{table:results}
\end{table}

Furthermore, to have a more accurate evaluation of our system, we used Free-Response Receiver Operating Characteristic (FROC) analysis \cite{kundel2008receiver}. Sensitivity at $7$ FP/scan rates (i.e. $0.125, 0.25, 0.5, 1, 2, 4, 8$) is computed and the corresponding results are plotted in Fig.~\ref{fig:dicecomparison}. The overall \textit{score} of system is defined as the average sensitivity for these $7$ FP/scan rates.  Our network achieved an average score of $\textbf{$\sim$92\%}$ (see Table.\ref{table:results2}).

\begin{table}[htb]
\caption{System performance in terms of sensitivity based on number of FPs/scan.}
\resizebox{\columnwidth}{!}{
\begin{tabular}{|>{\columncolor{gray}}l|c|c|c|c|c|c|c|l|}
\hline
FPs/scan & 0.125 & 0.25 & 0.5 & 1 & 2 & 4 & 8 & Average \\
\hline
Sensitivity & 0.773 & 0.870 & 0.924 & 0.941 & 0.962 & 0.980 & 0.986 & \textbf{0.919}  \\
\hline
\end{tabular}}
\label{table:results2}
\end{table}

\begin{figure}
\includegraphics[scale=0.37]{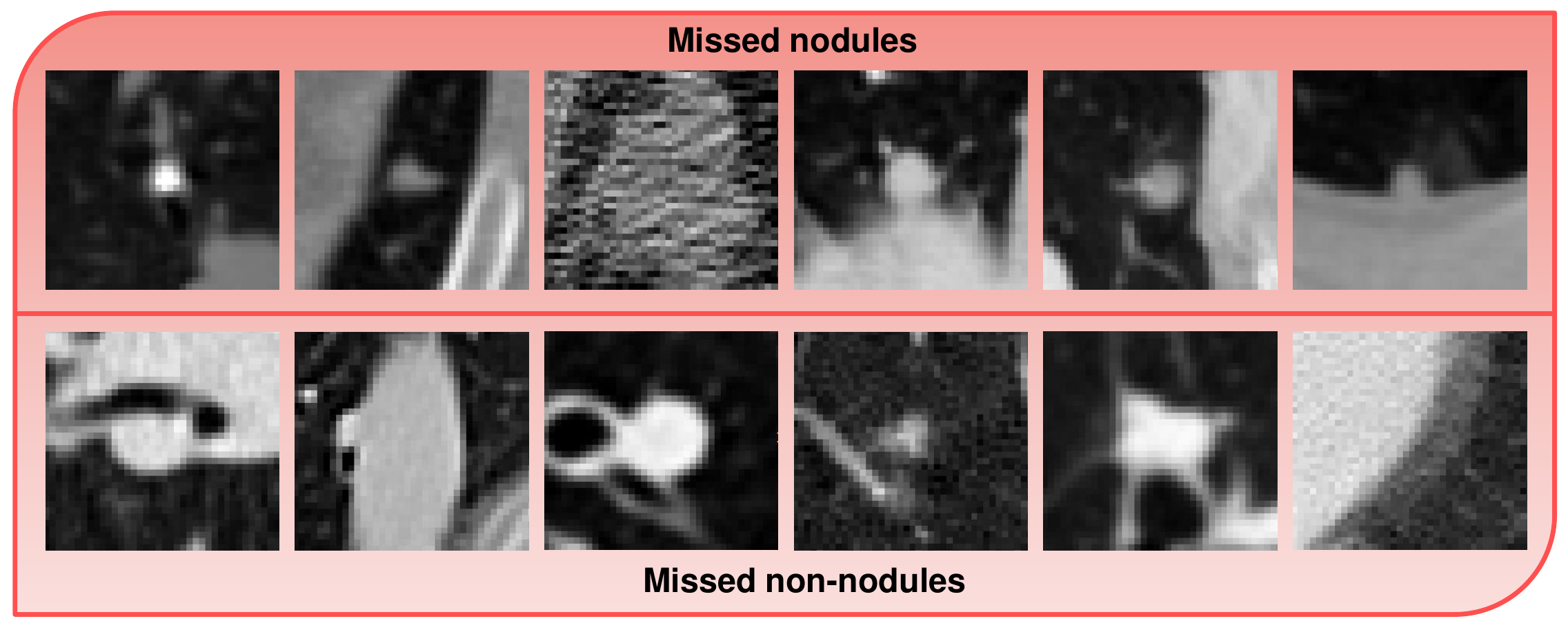}
\caption{Limitation of our system/failing cases: The first row shows $6$ examples of missing nodules. The bottom row shows some examples of non-nodules which are mistakenly considered as nodules.\label{fig:missingcases}}
\end{figure}
\section{Discussion and concluding remarks}
In this study, we proposed a 3D deep multi-task CNN for simultaneously performing segmentation and FP reduction. We showed that sharing some underlying features for these tasks and training a single model using shared features can improve the results for both tasks, which are critical for lung cancer screening. Furthermore, we showed that a semi-supervised approach can improve the results without the need for large number of labeled data in the training. It should be also note that there are some cases that our algorithm missed for FP reduction task. We illustrated some of those rarely seen examples of missing cases in Fig.\ref{fig:missingcases}. One reason seems to be the small size of the missed nodule. Alternatively, very similar appearance of missing cases to other abnormalities and normal lung parenchyma. As an alternative direction to semi-supervised approach, one may use GAN to generate realistic data. One recent study created realistic nodules to support this idea \cite{chuquicusma2017fool}.
\bibliographystyle{IEEEbib}
\bibliography{strings,refs}

\end{document}